\documentclass[runningheads]{llncs}
\usepackage[T1]{fontenc}
\usepackage{graphicx}
\usepackage{color}
\newcommand{\ourdata}[0]{HECSI}

\begin{document}
\title{ICPR 2024 Competition on\\ Multilingual Claim-Span Identification}
%
%

\author{
Soham Poddar\inst{1} \and 
Biswajit Paul\inst{2} \and 
Moumita Basu\inst{3} \and 
Saptarshi Ghosh\inst{1}
}

%
\institute{
Indian Institute of Technology Kharagpur, Kharagpur 721302, India \and
Centre for Artificial Intelligence and Robotics, DRDO, Bangalore 560093, India \and
NSHM Institute of Computing \& Analytics, NSHM Knowledge Campus, Kolkata 700053, India\\
}
\maketitle              
\begin{abstract}
A lot of claims are made in social media posts, which may contain misinformation or fake news. Hence, it is crucial to identify claims as a first step towards claim verification. Given the huge number of social media posts, the task of identifying claims needs to be automated. This competition deals with the task of `Claim Span Identification' in which, given a text, parts / spans that correspond to claims are to be identified. This task is more challenging than the traditional binary classification of text into claim or not-claim, and requires state-of-the-art methods in Pattern Recognition, Natural Language Processing and Machine Learning. For this competition, we used a newly developed dataset called \ourdata{} containing about 8K posts in English and about 8K posts in Hindi with claim-spans marked by human annotators. This paper gives an overview of the competition, and the solutions developed by the participating teams.
\end{abstract}

\section{Introduction}


A lot of people share alleged facts or `claims' on online social media. 
Toulmin~\cite{toulmin2003uses} defines a `claim' as ``\textit{a statement that
asserts something as true or valid, often without providing sufficient evidence for verification}''. 
Often, these alleged facts or claims contain unsupported or misleading information. Thus, it becomes crucial to detect such claims and fact-check them in order to prevent the spread of misinformation. 
Due to the sheer scale of social media, having millions of posts every day, 
automated methods are needed to identify claims for further processing and verification~\cite{gupta2021lesa,sundriyal2022empowering}. 
In this competition, we invited the participants to work on a  \textit{claim span identification} task, which we describe next.

\vspace{2mm}
\noindent \textbf{Claim Span Identification (CSI) Task:}
The \textit{Claim Span Identification}~(CSI) task is to extract which specific part(s) of a given text minimally represent a claim-like statement. Apart from filtering exact spans that need to be checked, these also serve as an explanation into why a sentence is labelled as a claim. Some examples of tweets and their claim-spans have been given in Table~\ref{tab:data_eg}.

Note that the CSI task is a more challenging and practically useful task than just a binary classification task of classifying a text as claim or not claim~\cite{gupta2021lesa}. While the binary classification is a sentence-level task, the CSI task is a word / token level task and requires a deeper understanding of the language semantics, including the broad pattern of how a claim may look like.

\begin{table}[!t]
    \centering
    \small
    \begin{tabular}{|p{116mm}|}
    \hline
     Don't understand why folks want to attend any sports event at this time. And \textcolor{blue}{\bf taking a vaccine developed in 365 days to me is high risk}. \\
     \hline
     You might want to read on side effects and also look how the \textcolor{blue}{\bf FDA has manipulated the stock market}. Look at Inovio, \textcolor{red}{\bf has a defense contract but no support for vaccine} \\
     \hline
     Not ready for this vaccine at all I am so nervous I hate shots and getting sick lol \\
     \hline
    \end{tabular}
    
    \vspace{1mm}
    \caption{Examples of input post text for the CSI task, from our dataset \ourdata. The target claim spans are highlighted (there is no claim in the last example).}
    \label{tab:data_eg}
\end{table}

\begin{table}[t]
    \centering
    \includegraphics[width=0.86\linewidth,trim={0 2.5mm 0 2.5mm},clip]{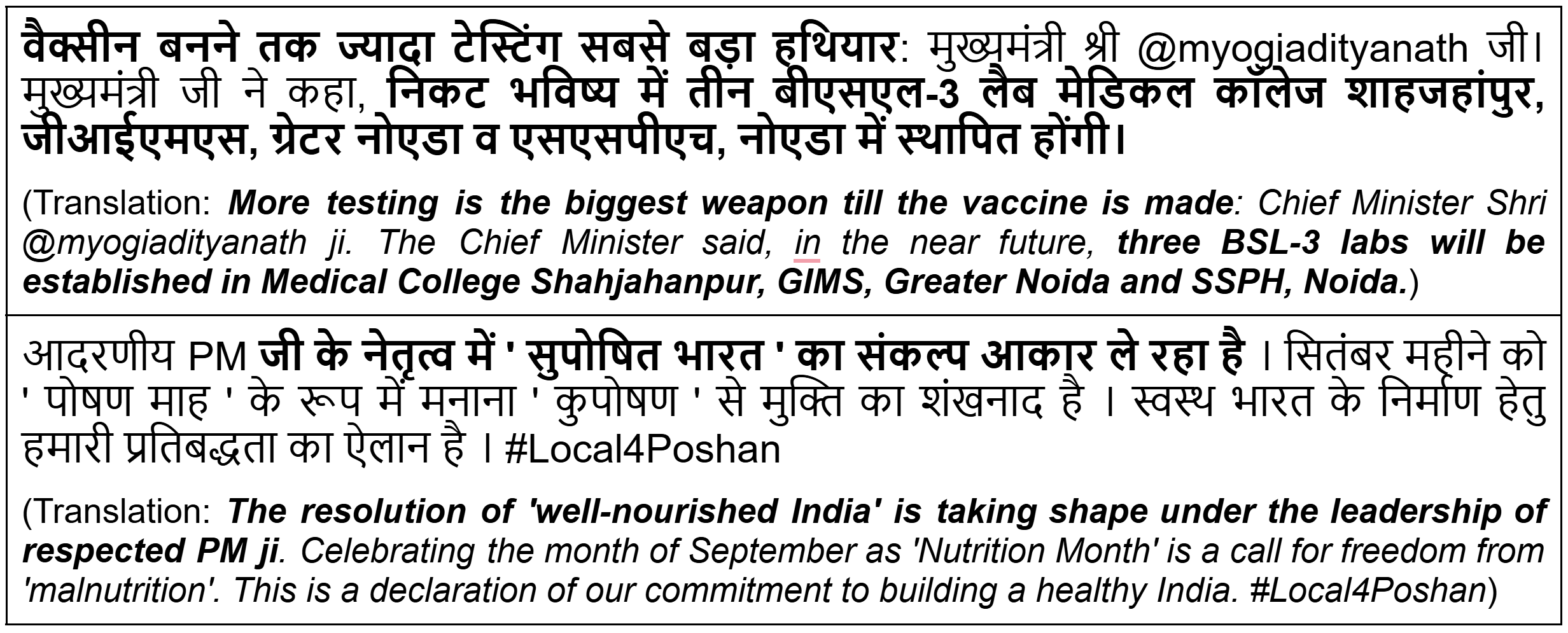}
    \caption{Examples of posts from \ourdata{} in Hindi (and their English translations) with claim spans highlighted. The claim spans have been marked by human annotators.}
    \label{tab:hindi-eg}
\end{table}

\vspace{2mm}
\noindent \textbf{Overview of the competition:} This competition\footnote{Competition website: \url{https://sites.google.com/view/icpr24-csi/home}} was organized as part of the ICPR 2024 conference\footnote{\url{https://icpr2024.org/}}. We invited participating teams to work on the claim-span identification task on a new dataset containing social media posts in two languages, English and Hindi, where claims are marked by human annotators. 
This paper gives an overview of the competition, including the dataset, and the solutions developed by the participating teams.

\section{Dataset for Claim-span Identification}
\label{sec:dataset}

\begin{table}[!t]
    \centering
    \small
    \begin{tabular}{|l|c|c|c|}
    \hline
    & \textbf{English (En)} &   \textbf{Hindi (Hi)} & \textbf{Multilingual (Ml)} \\
    \hline
    Total \#posts & 7,999 & 8,098 & 14,097 \\
    \#posts \textit{without claim spans} & 1,404 & 1,599 & 3,003 \\
    \#posts \textit{with a single claim span} & 4,837 & 5,568 & 10,405 \\
    \#posts \textit{with multiple claim spans} & 1,758 & 931 & 2,689 \\
    \hline
    \multicolumn{4}{|c|}{Splits for the competition} \\
    \hline
    Train set & 5,999 & 6,098 & 12,097* \\
    Validation set & 500 & 500 & 1,000* \\
    Test set & 1,500 & 1,500 & 3,000 \\
    \hline
    \end{tabular}
    \vspace{1mm}
    \caption{Statistics of the \ourdata{} dataset used for the competition. \textbf{*}~all kinds of data publicly available online was allowed for training models in the Multilingual track}.
    \label{tab:data_stat}
\end{table}

We developed a new dataset for the CSI task that (i)~contains claims that are important to detect in practice, (ii)~contains claims from different domains to improve the generalizability of the models trained on them, and (iii)~contains claims in two languages English and Hindi. 
We call the dataset \ourdata{} (\textbf{H}indi-\textbf{E}nglish \textbf{C}laim-\textbf{S}pan \textbf{I}dentification). 

For the English part of the \ourdata{}, we leveraged the CAVES dataset (from our prior work~\cite{poddar2022caves}) which contains \textit{anti-vaccine posts (tweets) about COVID19 vaccines}. Detecting anti-vaccine claims is important for understanding people's concerns about vaccines, in order to  
remove misconceptions / misinformation and improve adaptation of vaccines. 
For the Hindi part, we chose the CONSTRAINT dataset~\cite{bhardwaj2020hostility} which contains hostile (fake-news, hate-speech, etc.) social media posts in Hindi. Detecting fake-claims or hate-claims is important in order to counter such claims.

The \ourdata{} dataset was annotated by a team of human workers from a reputed annotation firm based in India\footnote{\url{https://www.cogitotech.com/}}. 
The annotators were selected such that they are fluent in English and Hindi, are familiar with Twitter and have substantial experience annotating tweets.
The annotators were also shown a few examples from existing claim datasets so that they could be acquainted with different types of claims. 
The annotators were asked to first judge whether a post contains a claim-like statement, and if yes, mark the minimal span(s) which represented the claim(s). The annotators were specifically instructed that \textit{all} claim-spans present in a post should be marked.
The annotation team was paid a mutually agreed compensation for their services.

The final dataset contains about $16k$ posts ($8k$ English and $8k$ Hindi). 
The dataset statistics are given in Table~\ref{tab:data_stat} (first 3 columns).
Examples from the final dataset are given in Table~\ref{tab:data_eg} for English and Table~\ref{tab:hindi-eg} for Hindi. 
Additionally, we created a Multilingual dataset by combining the English and Hindi datasets, as shown in Table~\ref{tab:data_stat} (last column).

Importantly, \ourdata{} contains several \textit{posts with multiple claim spans} and \textit{posts without any claim spans} (see Table~\ref{tab:data_stat}), which makes the CSI task particularly more challenging on this dataset.

\vspace{1mm}
\noindent \textbf{Dataset availability:} 
We make the HECSI dataset publicly available for promoting further research on the important problems of claim-span identification and claim detections.\footnote{The dataset is available at \url{https://github.com/sohampoddar26/hecsi-data}}

\vspace{1mm}
\noindent \textbf{Data splits for the competition:} 
For the competition, we  randomly split each of the English and Hindi datasets into validation sets~($500$ samples in each language), test sets~($1,500$ samples in each language) and training sets (remaining samples). Table~\ref{tab:data_stat} states the statistics of the train, validation and test sets used in the competition. 
The participants were evaluated based on the performance of their models on the three test sets -- 
separately for the English test set and the Hindi test set, and the Multilingual test set (obtained by combining the English and Hindi test sets).

\section{Competition Tracks and Evaluation}

There were three different tracks for the competition. The three tracks are as follows.

(1) \textbf{Constrained English track} -- In this track, only the English training and validation sets provided for this competition (as described in the previous section) could be used to train / fine-tune models, and only the \textit{English test set} was used for evaluation of models. Teams were forbidden from using any other data for training / fine-tuning models. 

(2) \textbf{Constrained Hindi track} -- In this track, only the Hindi training and validation sets provided for this competition could be used to train / fine-tune models, and only the \textit{Hindi test set} was used for evaluation. Teams were forbidden from using any other data for training / fine-tuning models. 

(3) \textbf{Unconstrained Multilingual track} -- Here the teams were free to use any kind of resources to train / fine-tune models, including both the English and Hindi train sets provided for this competition and any external data. 
A combination of the English and Hindi test sets, containing 1,500 English posts and 1,500 Hindi posts, was used for evaluation of models in this track (see Table~\ref{tab:data_stat}, last column).

Out of the three tracks, a team could participate in one or more tracks. For each track, a team was allowed to submit up to 2 `runs' or solutions, so that they could try out two methods.

\vspace{2mm}
\noindent \textbf{Evaluation:}  
For evaluation, we followed the strategy of Sundriyal et.al.~\cite{sundriyal2022empowering}, which introduced the CSI task. 
We treated the CSI task as a \textit{binary token-classification} task, where each text token has to be predicted as part of a claim~(class $1$) or not~(class $0$).
Thus the prediction for a given post was expected to be a binary vector (consisting of 0s and 1s) of length equal to the number of tokens in the post. 
Note that submissions included token predictions at the word level and \textit{not} at the tokenized sub-word level (as in transformer-based models).

For evaluation of this token-level classification, we used the standard metrics \textit{Macro-F1} score\footnote{\url{https://scikit-learn.org/stable/modules/generated/sklearn.metrics.f1\_score.html}} and \textit{Jaccard score}\footnote{\url{https://scikit-learn.org/stable/modules/generated/sklearn.metrics.jaccard\_score.html}}, which were calculated at the token level. 
Given binary vectors of gold-standard and prediction labels for a particular data point (text), the metrics \textit{Macro-F1~(M-F1)} and \textit{Jaccard~(Jacc)} are calculated as follows:
\[\texttt{M-F1} = \frac{TP}{2 \cdot TP + FP + FN} +  \frac{TN}{2 \cdot TN + FP + FN} \]
\[\texttt{Jacc} = \frac{TP}{TP + FP + FN} \]
where TP, TN, FP, FN carry their usual meanings of True Positive, True Negative and False Positive, False Negative respectively (\textbf{1} is the positive class).

For each input text, the above metrics was calculated over all the tokens in the text. Then these metrics were averaged over all the posts in the test set for a given track.\footnote{Participating teams were provided with an evaluation script computing the metrics.} 

\begin{table}[!t]	
\centering
\small
\begin{tabular}{|l|p{0.78\columnwidth}| } 
\hline
\textbf{Team name}& \textbf{Institute(s) of the team members}  \\
\hline
 
ClaimCatchers & Amrita School of Artificial Intelligence, Coimbatore, Amrita Vishwa Vidyapeetham, India \\ \hline

DLRG & Vellore Institute of Technology Chennai, India \\ \hline

FactFinders & Indraprastha Institute of Information Technology, Delhi, and IIIT Dharwad, India \\\hline

GMU-MU & George Mason University, USA \\ \hline

JU\_NLP &  Jadavpur University, IIEST Shibpur, and RCCIIT,   India \\ \hline

KSK& Novartis, Hyderabad and MiQ, Bangalore, India \\ \hline

neural\_nexus & Amity University Kolkata, India \\  \hline

NLP4Good & Kerala University of Digital Sciences, Innovation and Technology, Thiruvananthapuram, India.\\ \hline

RateLimit\_ Error & City University, Bangladesh \\ \hline

\end{tabular}
\vspace{2mm}
\caption{Details of the participating teams}
\label{tab:teamdetails}
\end{table}

\section{Participating Teams and Results}


\noindent \textbf{Participating teams:} Nine (9) teams submitted their solutions to the tasks.\footnote{A few other teams took the data at the beginning of the competition, but finally did not submit solutions.} 
The details of these nine teams 
are reported in Table~\ref{tab:teamdetails}. 
While most teams were from India, there was participation from USA and Bangladesh as well.

\vspace{2mm}
\noindent \textbf{Organizer baseline:} Apart from the participating teams, the organizers of this competition also provided a baseline model.
This was a simple bert-base-multilingual-uncased model, paired with a linear token classification layer. The only preprocessing done to the data was to replace the URLs with the tag `HTTPURL'.
A batch size of $128$ and learning rate of $10^{-5}$ was used for fine-tuning the model for $5$ epochs.
During inference, we considered a word token to be part of a claim if any of its tokenized sub-word tokens were predicted as $1$.

This baseline model was trained over only the training sets given for this competition (without using any external resources), and evaluated over the same test set (for each track) as the runs submitted by the participating teams.

\vspace{2mm}
\noindent \textbf{Results:}
Table~\ref{tab:results} shows the performances of all runs submitted by all the participating teams, and the organizer baseline model. The metric scores for each track are averaged over all samples in the test set of the corresponding track, and are reported as percentages. 
The JU\_NLP team performed the best in the Constrained English and Hindi tracks, while FactFinders and DLRG obtained best scores in the Unconstrained Multilingual track.

For the Constrained English and Hindi tracks, the JU\_NLP team with their fine-tuned multilingual language model (XLM-RoBERTa) performed the best in both metrics with M-F1 of \textbf{74.8}, Jaccard of \textbf{54.5} for English, and M-F1 of \textbf{81.7}, Jaccard of \textbf{67.1} for Hindi, thus easily beating the organizer baseline scores. 
The FactFinders team only slightly beat the organizer baseline in both English and Hindi tracks in terms of macro-F1, 
with DLRG team performing close to it, beating the baseline only in Jaccard score in the English track.
The GMU-MU team achieved the second-highest Jaccard score (52.1) in the English track, where they outperformed the organizer baseline.

Only $5$ teams submitted runs for the unconstrained multilingual track, with DLRG performing the best in Macro-F1 scores~(\textbf{59.3}), and FactFinders performing the best in Jaccard score~(\textbf{41.4}).
However, neither of them could beat the organizer baseline scores for this track.

\begin{table}[tb]
    \centering
    \def\arraystretch{1.3}
    \setlength\tabcolsep{4pt}
    \begin{tabular}{|l|c||c|c||c|c||c|c|}
    \hline
        \textbf{Team Name} & \textbf{Run}   & \multicolumn{2}{c||}{\bf English} & \multicolumn{2}{c||}{\bf Hindi} & \multicolumn{2}{c|}{\bf Multilingual}	\\
        \cline{3-8}
		         &   \#  & \textbf{M-F1} & \textbf{Jacc} & \textbf{M-F1} & \textbf{Jacc} & \textbf{M-F1} & \textbf{Jacc} \\
           
        \hline \hline
   Organizer baseline & 1 & 72.6 & 50.3 & 74.2 & 63.6 & 74.2 & 57.4 \\
        \hline \hline
        ClaimCatchers & 1 & 46.1 &  5.0 & 38.8 & 47.3 & 42.4 & 26.1 \\
        	       & 2 & 40.9 & 21.9 & 39.4 & 23.7 & 40.2 & 22.8 \\

        \hline
        DLRG          & 1 & 71.2 & 50.8 & 61.0 & 53.1 & \textbf{59.3} & 37.8 \\
        
        \hline
        FactFinders   & 1 & 72.8 & 50.3 & 75.9 & 63.6 & 55.8 & \textbf{41.4} \\
        	       & 2 & 72.2 & 50.3 & 72.9 & 62.0 &    - &    - \\

        \hline
        GMU-MU        & 1 & 67.3 & 52.1 & 68.0 & 62.4 &    - &    - \\
        	       & 2 & 64.8 & 50.4 &    - &    - &    - &    - \\

        \hline
        JU\_NLP       & 1 & \textbf{74.8} & \textbf{54.5} & \textbf{81.7} & \textbf{67.1} &    - &    - \\
        	       & 2 &    - &   -  & 78.4 & 63.1 &    - &    - \\

        \hline
        KSK           & 1 & 44.0 &  0.9 & 36.2 &  2.4 & 39.6 &  0.9 \\
        	       & 2 & 44.0 &  0.9 & 36.2 &  2.4 & 39.6 &  0.9 \\

        \hline
        neural\_nexus & 1 & 38.2 & 30.6 & 37.5 & 31.0 & 37.9 & 30.8 \\

        \hline
        NLP4Good	  & 1 & 68.1 & 43.8 &    - &    - &    - &    - \\

        \hline
     RateLimit\_Error & 1 & 27.9 & 38.4 & 37.4 & 56.3 &    - &    - \\

        \hline
    \end{tabular}
    \vspace{2mm}
    \caption{Performance of the participating teams and the organizer baseline (Macro-F1 and Jaccard reported as percentages). Note that some of the teams participated in one or two tracks only (out of the three tracks English, Hindi, Multilingual). For each track, a team was allowed to submit up to 2 runs, so that they could try out two methods. The best score of the participants for every metric is highlighted in boldface.}
    \label{tab:results}
\end{table}

\subsection{Solutions submitted by the teams}

In this section, we describe the solutions adopted by the top three teams, which are JU\_NLP, FactFinders and DLRG, followed by a brief overview of the methods applied by other teams.

\vspace{2mm}
\noindent\textbf{JU\_NLP:}
This team applied several pre-processing steps over the data, including the removal of escape characters, standardization of URLs, substitution of user mentions with `@user', replacement of unrecognized Unicode tokens with `unk', and case folding of tokens to lowercase. After pre-processing, the text was tokenized and used to train two pre-trained transformer based models BERT-Base-Multilingual-Cased~\cite{devlin2018bert} and XLM-RoBERTa-Base~\cite{conneau2019unsupervised}. 
Both models, which support multiple languages including English and Hindi, were provided with token IDs, attention masks, and claim labels to predict the correct claim tags corresponding to the claim spans in the text.
The models were fine-tuned using the following hyperparameters -- Optimizer: AdamW, Learning Rate: 2e-5, Batch Size: 8, Epochs: 3, Weight Decay: 0.01.
They achieved the best performance in both the English and Hindi constrained tracks, and did not participate in the Multilingual track.

\vspace{2mm}
\noindent \textbf{FactFinders:} This team fine-tuned various pre-trained transformer-based models. 
For the English track, they employed fine-tuned DeBERTa-Large~\cite{he2021debertav3}, Distil-BERT~\cite{sanh2019distilbert}, and RoBERTa-base models~\cite{liu2019roberta}. 
For the Hindi track, they fine-tuned multilingual models such as Muril-Base~\cite{khanuja2021muril}, XLM-RoBERTa-Large~\cite{conneau2019unsupervised}, and Multilingual BERT~\cite{devlin2018bert}. 
Finally, for the unconstrained Multilingual track, they leveraged XCLAIM~\cite{mittal2023lost} and CURT~\cite{sundriyal2022empowering} datasets along with \ourdata{} to fine-tuned DeBERTa-Large model. 
They also modified the default pre-trained language models (PLMs) with two approaches--\\
(i)~They introduced the Cosine Attention Integration (CAI), which enhances encoder models by adding a Cosine Similarity (between each token and the entire sequence) layer along with another Multi-Head Attention layer. Then, these are added and passed to the linear classification layer.\\
(ii)~In the second approach, the Ensemble Majority Voting technique is applied, which enhances overall accuracy by combining predictions from multiple models and selecting the most common prediction as the final output.

Among the models they tried, the DeBERTa-Large model with CAI was identified as the best-performing model for both the English and Multilingual datasets. Whereas, the Muril-Base model with CAI achieved the best performance on the Hindi dataset.
This team achieved the best Jaccard score and the second-best Macro-F1 score in the Multilingual track, and second best in the English and Hindi tracks.

\vspace{2mm}
\noindent \textbf{DLRG:} This team used 
PLMs, such as XLM-RoBERTa, mBERT, or mDEBERTa along with token classification layer into a 3-class BIO scheme instead of the \ourdata's IO scheme.
They also tried the DABERTA~\cite{sundriyal2022empowering} model.
The XLM-RoBERTa model performed the best, which was trained with the following hyperparameters -- a batch size of 1, 40 epochs, and a learning rate of 1e-5 for the PLM. 
For the unconstrained multilingual track, they also trained their models over the XCLAIM dataset.
This team achieved the best Macro-F1 score in the Multilingual track. 

\vspace{2mm}
\noindent\textbf{GMU\_MU:}
This team instruction-tuned different LLMs using Alpaca-style prompts -- Mistral-7B-v0.3~\cite{jiang2023mistral}, Llama-3-8B~\cite{meta2024introducing} and Bloomz-7b1~\cite{muennighoff2022crosslingual}. The prompt consists of a task-specific instruction. 
The LoRA (Low-Rank Adaptation)~\cite{hu2021lora} method was employed for fine-tuning, with parameters set to $r$ = 64, $\alpha$ = 16.
Other hyperparameters: 10 max steps, a per-device batch size of 16, 2 gradient accumulation steps, the paged\_adamq\_32bit optimizer, a 2e-4 learning rate, a 0.1 warmup ratio, a cosine scheduler, and a maximum sequence length of 1024.  
The use of LLMs with instruction fine-tuning using LoRA showed encouraging performance in terms of Jaccard in the English track (where this team achieved the second-highest score, outperforming the organizer baseline).

\vspace{2mm}
\noindent \textbf{ClaimCatchers:} They used BERT-large-uncased~\cite{devlin2018bert} and IndicBERT~\cite{kakwani2020indicnlpsuite} as base models due to their suitability for token classification. 
IndicBERT is a multilingual model pre-trained on 12 major Indian languages. 
Both models were fine-tuned with three output labels (BIO scheme), using 10 epochs, a batch size of 8, and a learning rate of 2e-5, with mixed precision training for faster results.

\vspace{2mm}
\noindent \textbf{KSK:} This team mainly employed XLM-RoBERTa for English, Hindi, and multilingual datasets
and MuRIL specifically for the Hindi dataset. Both of these models are optimized for strong multilingual and Indian language processing. The team configured hyperparameters as follows: CosineAnnealingLR scheduler with T\_max and T\_0 set to 10, a minimum learning rate of 1e-6, and a maximum learning rate of 2e-5. They fine-tuned the models for 10 epochs with a weight decay of 1e-4 and used a seed of 42.

\vspace{2mm}
\noindent \textbf{neural\_nexus:} They trained a multilingual token classification model utilizing RoBERTa. They incorporated various pre-processing techniques and employed the LoRA (Low-Rank Adaptation)~\cite{hu2021lora} method for efficient fine-tuning, with the following settings: Rank ($r$) of 8, $\alpha$ of 32, and a dropout rate of 0.1. 
Other training parameters included a batch size of 8, a learning rate of 5e-5, 3 epochs, and 500 warm-up steps. AdamW was used as the optimizer, with a linear learning rate scheduler to adjust the learning rate during training.

\vspace{2mm}
\noindent \textbf{NLP4Good:} This team used BERT to classify tokens in text as part of a claim or not. Their key steps include tokenizing text, generating labels for claims, and fine-tuning a pretrained BERT, optimizing with AdamW.

\vspace{2mm}
\noindent \textbf{RateLimit\_Error:} They employed Logistic Regression as their classifier and used SMOTE~\cite{chawla2002smote} within a pipeline to address class imbalance. They considered 10\% of the training dataset as a validation set with a seed of 42.

\subsection{Analysis of the results}

We now present some observations on the techniques used and the results obtained by the teams. 
We noticed that fine-tuned transformer models such as BERT, RoBERTa, DeBERTa, XLM-RoBERTa, and MuRIL were predominantly used by most teams. 
Team KSK utilized the CosineAnnealingLR scheduler to adjust the learning rate during training. 
Meanwhile, the FactFinder team introduced the Cosine Attention Integration (CAI) technique, a novel approach that enhances encoder models by combining multi-head attention with a cosine similarity layer, to improve model performance.

Proper pre-processing with suitable replacement of non-informative tokens potentially has a positive impact on performance, as observed from the steps taken by the  JU\_NLP team 
who achieved top performance in the English and Hindi tracks by fine-tuning BERT-Base-Multilingual-Cased and XLM-RoBERTa-Base models 
(as shown by the results in Table~\ref{tab:results}). 
LLMs also perform well for the English track  as observed from GMU-MU submission, but are not quite as good as fine-tuned discriminative Transformer-based models.

Only five teams participated in the Unconstrained Multilingual track, and the performances of the classifiers in this track was notably lower.
The highest Macro-F1 score in the Unconstrained Multilingual track was \textbf{59.3}, achieved by team \textit{DLRG}, and the highest Jaccard score was \textbf{41.4}, achieved by team \textit{FactFinders}, as shown in Table~\ref{tab:results}. In contrast, the highest Macro-F1 and Jaccard scores in the Constrained English and Hindi tracks were considerably higher.
The results indicate that the Unconstrained Multilingual track posed a significantly greater challenge for participants in terms of developing an efficient classifier, even though the top-performing teams leveraged additional CSI datasets (like XCLAIM and CURT) to train their models.
It can also be noted that the organizer baseline method in the Unconstrained Multilingual track, trained on only the English and Hindi HECSI train sets, outperformed all participating teams in both Macro-F1 (\textbf{74.2}) and Jaccard scores (\textbf{57.4}).
The fact that use of additional CSI data along with HECSI failed to yield positive results for multilingual track, as observed from the DLRG and FactFinders submissions, perhaps indicates the need for improved domain augmentation strategies for better transfer learning.

\section{Conclusion}

We organized the ICPR 2024 Competition on Multilingual Claim-Span Identification, which is a crucial step towards claim verification, and is much more challenging than standard binary classification of claims vs. non-claims. 
For this competition, we developed the \ourdata{} dataset consisting of posts in Hindi and English, and participants were asked to identify spans of text which represent claims, from these posts. 
Though the participants employed a variety of diverse techniques, no team could achieve significantly better results than the baseline, indicating that the Claim Span Identification (CSI) task and the \ourdata{} dataset still pose considerable challenges, and the scope for research is wide open.
We make the \ourdata{} publicly available (see Section~\ref{sec:dataset}) to promote further research on the important problem of claim-span identification in various languages.

\section*{Acknowledgements}

We thank the ICPR 2024 organizers for the opportunity of running this competition, and their support during the event. We also thank all the participants for their interest in this competition. 
Development of the dataset for the competition was supported by a research grant from Defence Research and Development Organisation (DRDO), Government of India. 
The first author (S. Poddar) is supported by the Prime Minister’s Research Fellowship (PMRF) from the Ministry of Education, Government of India.

%
%

\bibliographystyle{splncs04}
\bibliography{ref}

\begin{thebibliography}{10}
\providecommand{\url}[1]{\texttt{#1}}
\providecommand{\urlprefix}{URL }
\providecommand{\doi}[1]{https://doi.org/#1}

\bibitem{bhardwaj2020hostility}
Bhardwaj, M., Akhtar, M.S., Ekbal, A., Das, A., Chakraborty, T.: Hostility detection dataset in hindi. arXiv preprint arXiv:2011.03588  (2020)

\bibitem{chawla2002smote}
Chawla, N.V., Bowyer, K.W., Hall, L.O., Kegelmeyer, W.P.: Smote: synthetic minority over-sampling technique. Journal of artificial intelligence research  \textbf{16},  321--357 (2002)

\bibitem{conneau2019unsupervised}
Conneau, A., Khandelwal, K., Goyal, N., Chaudhary, V., Wenzek, G., Guzm{\'a}n, F., Grave, E., Ott, M., Zettlemoyer, L., Stoyanov, V.: Unsupervised cross-lingual representation learning at scale. arXiv preprint arXiv:1911.02116  (2019)

\bibitem{devlin2018bert}
Devlin, J., Chang, M.W., Lee, K., Toutanova, K.: Bert: Pre-training of deep bidirectional transformers for language understanding. arXiv preprint arXiv:1810.04805  (2018)

\bibitem{gupta2021lesa}
Gupta, S., Singh, P., Sundriyal, M., Akhtar, M.S., Chakraborty, T.: Lesa: Linguistic encapsulation and semantic amalgamation based generalised claim detection from online content. arXiv preprint arXiv:2101.11891  (2021)

\bibitem{he2021debertav3}
He, P., Gao, J., Chen, W.: Debertav3: Improving deberta using electra-style pre-training with gradient-disentangled embedding sharing. arXiv preprint arXiv:2111.09543  (2021)

\bibitem{hu2021lora}
Hu, E.J., Shen, Y., Wallis, P., Allen-Zhu, Z., Li, Y., Wang, S., Wang, L., Chen, W.: Lora: Low-rank adaptation of large language models. arXiv preprint arXiv:2106.09685  (2021)

\bibitem{jiang2023mistral}
Jiang, A., Sablayrolles, A., Mensch, A., Bamford, C., Chaplot, D., de~las Casas, D., Bressand, F., Lengyel, G., Lample, G., Saulnier, L., et~al.: Mistral 7b (2023). arXiv preprint arXiv:2310.06825  (2023)

\bibitem{kakwani2020indicnlpsuite}
Kakwani, D., Kunchukuttan, A., Golla, S., Gokul, N., Bhattacharyya, A., Khapra, M.M., Kumar, P.: Indicnlpsuite: Monolingual corpora, evaluation benchmarks and pre-trained multilingual language models for indian languages. In: Findings of the Association for Computational Linguistics: EMNLP 2020. pp. 4948--4961 (2020)

\bibitem{khanuja2021muril}
Khanuja, S., Bansal, D., Mehtani, S., Khosla, S., Dey, A., Gopalan, B., Margam, D.K., Aggarwal, P., Nagipogu, R.T., Dave, S., et~al.: Muril: Multilingual representations for indian languages. arXiv preprint arXiv:2103.10730  (2021)

\bibitem{liu2019roberta}
Liu, Y., Ott, M., Goyal, N., Du, J., Joshi, M., Chen, D., Levy, O., Lewis, M., Zettlemoyer, L., Stoyanov, V.: Roberta: A robustly optimized bert pretraining approach. arXiv preprint arXiv:1907.11692  (2019)

\bibitem{meta2024introducing}
Meta, A.: Introducing meta llama 3: The most capable openly available llm to date. Meta AI  (2024)

\bibitem{mittal2023lost}
Mittal, S., Sundriyal, M., Nakov, P.: Lost in translation, found in spans: Identifying claims in multilingual social media. In: Proceedings of the 2023 Conference on Empirical Methods in Natural Language Processing. pp. 3887--3902 (2023)

\bibitem{muennighoff2022crosslingual}
Muennighoff, N., Wang, T., Sutawika, L., Roberts, A., Biderman, S., Scao, T.L., Bari, M.S., Shen, S., Yong, Z.X., Schoelkopf, H., et~al.: Crosslingual generalization through multitask finetuning. arXiv preprint arXiv:2211.01786  (2022)

\bibitem{poddar2022caves}
Poddar, S., Samad, A.M., Mukherjee, R., Ganguly, N., Ghosh, S.: Caves: A dataset to facilitate explainable classification and summarization of concerns towards covid vaccines. In: Proceedings of the 45th International ACM SIGIR Conference on Research and Development in Information Retrieval. pp. 3154--3164 (2022)

\bibitem{sanh2019distilbert}
Sanh, V., Debut, L., Chaumond, J., Wolf, T.: Distilbert, a distilled version of bert: smaller, faster, cheaper and lighter. arXiv preprint arXiv:1910.01108  (2019)

\bibitem{sundriyal2022empowering}
Sundriyal, M., Kulkarni, A., Pulastya, V., Akhtar, M.S., Chakraborty, T.: Empowering the fact-checkers! automatic identification of claim spans on twitter. In: Proceedings of the 2022 Conference on Empirical Methods in Natural Language Processing. pp. 7701--7715 (2022)

\bibitem{toulmin2003uses}
Toulmin, S.E.: The uses of argument. Cambridge university press (2003)

\end{thebibliography}

\end{document}